\documentclass[journal,comsoc]{IEEEtran}
\usepackage{amsmath,amsfonts}
\usepackage{algorithmic}
\usepackage{algorithm}
\usepackage{array}
\usepackage[caption=false,font=normalsize,labelfont=sf,textfont=sf]{subfig}
\usepackage{textcomp}
\usepackage{stfloats}
\usepackage{url}
\usepackage{verbatim}
\usepackage{graphicx}
\usepackage{cite}
\usepackage[utf8]{inputenc}
\usepackage[english]{babel}
\usepackage{multirow}
\usepackage[backref]{hyperref}

\begin{document}

\title{Semantic Feature Decomposition based Semantic Communication System of Images with Large-scale Visual Generation Models}

\author{Senran Fan, Zhicheng Bao, Chen Dong*, Haotai Liang, Xiaodong Xu, \textit{Senior Member}, \textit{IEEE},  Ping Zhang, \textit{Fellow}, \textit{IEEE}
\thanks{Senran Fan, Zhicheng Bao, Haotai Liang, Xiaodong Xu and Ping Zhang are with the State Key Laboratory of Networking and Switching Technology, Beijing University of Posts and Telecommunications, Beijing, 100876, China. (E-mail: FSR@bupt.edu.cn; zhicheng$\_$bao@bupt.edu.cn; lianghaotai@bupt.edu.cn; xuxiaodong@bupt.edu.cn; pzhang@bupt.edu.cn)}
\thanks{*Chen Dong is the corresponding author and with the State Key Laboratory of Networking and Switching Technology, Beijing University of Posts and Telecommunications, Beijing, 100876, China. (E-mail: dongchen@bupt.edu.cn)}
\thanks{X. Xu and P. Zhang are also with the Department of Broadband Communication, Peng Cheng Laboratory, Shenzhen, China.}}

\markboth{}
{}

\maketitle

\begin{abstract}
    The end-to-end image communication system has been widely studied in the academic community. The escalating demands on image communication systems in terms of data volume, environmental complexity, and task precision require enhanced communication efficiency, anti-noise ability and semantic fidelity. Therefore, we proposed a novel paradigm based on Semantic Feature Decomposition (SeFD) for the integration of semantic communication and large-scale visual generation models to achieve high-performance, highly interpretable and controllable image communication. According to this paradigm, a Texture-Color based Semantic Communication system of Images—TCSCI is proposed. TCSCI decomposing the images into their natural language description (text), texture and color semantic features at the transmitter. During the transmission, features are transmitted over the wireless channel, and at the receiver, a large-scale visual generation model is utilized to restore the image through received features. TCSCI can achieve extremely compressed, highly noise-resistant, and visually similar image semantic communication, while ensuring the interpretability and editability of the transmission process. The experiments demonstrate that the TCSCI outperforms traditional image communication systems and existing semantic communication systems under extreme compression with good anti-noise performance and interpretability.  
\end{abstract}

\begin{IEEEkeywords}
    Semantic Communication, Large-scale Visual Generation Models, SeFD, TCSCI
\end{IEEEkeywords}

\section{Introduction}
\IEEEPARstart{T}he way humans understand and process information is frequently grounded in the decomposition of semantic features. For audio, this could involve breaking down the signal into distinct elements like timbre, pitch or loudness. Similarly, for images, the decomposition often involves segmenting the visual information into meaningful components such as colors, textures, edges, and depths. This process of semantic decomposition allows the human brain to extract and interpret the underlying structure and meaning from complex sensory inputs.\par
This provides inspiration that when designing end-to-end communication systems, we can also decompose the information source into different semantic components for transmission, and perform semantic synthesis and information reconstruction at the receiver. By drawing on the human perception process of information, this new communication paradigm may possess higher compression ratio, stronger explainability, controllability, and editability.\par
And in fields of image communication, end-to-end image communication systems cover the entire process of image compression, encoding, transmission and restoration. With the development of technologies like Internet of Things, the communication volume, complexity of the communication environment and various requirements of the communication tasks and diverse communication task requirements are constantly growing. This has driven more research into end-to-end image communication systems and has led to three key requirements being placed on these systems:
\begin{enumerate}
    \item \textbf{Communication Efficiency}, that is, whether the system can be applied to tasks with limited channel resources.
    \item \textbf{Anti-noise Ability}, that is, whether the system's anti-noise performance can support normal transmission in complex communication scenarios, such as low signal-to-noise ratio conditions.
    \item \textbf{Semantic Fidelity}, that is, whether the image transmission system can protect the key semantic content of the task focus in the image, in order to achieve normal task execution.
\end{enumerate}  \par
Traditional image communication systems, based on symbol-level error-free transmission, have reached close to the Shannon limit in terms of communication efficiency at the physical layer, making it difficult to achieve significant further improvement. On the other hand, their transmission is based on source-channel separation coding pattern, and under low signal-to-noise ratio conditions, the noise-resistance performance of traditional systems exhibits a "cliff effect", making it difficult for the system to adapt to complex and harsh communication scenarios. To solve this, researchers in the field of communications have been constantly exploring new communication theories and paradigms, and the concept of semantic communication has received increasing attention and exploration. First proposed in Shannon's paper \cite{shannon}, semantic communication systems extract, transmit and reconstruct the semantic information of the source. Accoring to \cite{6G,semantic}, by relaxing the requirement for error-free information transmission, semantic communication technology focuses on task-relevant semantic information, removes redundant data to the task, reduces the amount of transmitted data, and improves transmission efficiency. It also introduces joint source-channel coding technology \cite{jscc0, jscc} to replace the traditional independent source and channel coding, avoiding the cliff effect of channel coding under low signal-to-noise ratio conditions, and has stronger anti-noise ability in complex and harsh communication environments.\par
At the same time, with the continuous development of Artificial Intelligence and computing power, large-scale visual models have achieved remarkable results in the fields of image recognition \cite{vit}, object detection, semantic segmentation \cite{sam}, image compression, and image generation \cite{latent}. Particularly in image compression and generation, the emergence of large-scale models like SD (Stable Diffusion) for image generation has made it possible to achieve higher compression ratios for images and semantically-controlled image restoration. The "Text+Sketch" image compression algorithm proposed in paper \cite{textsketch} extracts the edge features of the image and uses ControlNet \cite{controlnet} to control the SD model for image restoration, achieving semantically-controlled extreme image compression. However, the restored images after decompression have a relatively large visual difference from the original images. Inspired by this, combining large-scale image generation models with semantic communication systems holds the promise of improving the compression ratio at the transmitter and achieving higher-quality semantic reconstruction at the receiver. It can also bring stronger controllability and interpretability to the semantic extraction and reconstruction process. \par

\begin{table*}[ht]
    \renewcommand\arraystretch{1.7}
    \begin{flushleft}   
    \caption{Comparison between Image Communication Systems} 
	\centering
    \label{table:1}
    \begin{tabular}{|m{1.2cm}|m{2.0cm}|m{3.8cm}|m{2.0cm}|m{2.0cm}|m{2.0cm}|m{2.0cm}|}
        \hline Type & System & Transmission Content & Compression Ratio & Anti-noise Ability & Visual Similarity & Interpretability \\ 
        \hline Tradition & —— & Encoding encoded by Entropy Coding and Channel Coding & Low & Low & High & High \\
        \hline \multirow{3}*{Semantic} & Semantic Feature-driven & Encoding compressed from Explicit Semantic Features defined by Human or Task & High & High & Low & High \\ 
        \cline{2-7} & Model-driven & Encoding encoded by Black-box End-to-end Neural Network & Medium & High & High & Low \\
        \cline{2-7} & TCSCI & Encoding compressed from Text, Texture and Color Semantic Features & High & High & High & High \\ 
        \hline
    \end{tabular}
\end{flushleft} 
\end{table*}

Accoring to Table I, the existing research on image semantic communication can be mainly divided into two categories: semantic feature-driven \cite{optimization} and model-driven \cite{lsci, sci1, sci2, stsci}. The semantic feature-driven image semantic communication system extracts targeted semantic features from the image for transmission and reconstruction, which extracts the textual description from the image and uses a text-driven image generation model at the receiver to generate the image. The model-driven system, on the other hand, is based on the autoencoder architecture, using end-to-end trained convolutional neural networks to compress the high-dimensional image information into low-dimensional semantic vectors at the transmitter, and then reconstruct the image at the receiver. The advantage of the semantic feature-driven approach is that it has a high compression ratio and strong interpretability, but since the selected semantic features cannot cover all the information in the image, the generated image at the receiver has a large visual difference from the original image, and can only be applied to limited tasks. The compression ratio of the model-driven approach exceeds traditional image compression algorithms, but is relatively lower than the semantic feature-driven approach. The reconstructed images are more visually similar to the original, but the process of semantic information extraction, compression, and reconstruction is performed in a neural network-like black box, resulting in poor controllability and interpretability.\par
Given the background needs and the advantages and limitations of the existing work, it's hoped that the proposed system will have the following characteristics:
\begin{enumerate}
    \item Ability to serve as a feasible example of the novel paradigm for the integration of semantic communication and large-scale visual generation models, providing insights for future research in this feild. 
    \item Ability to achieve a high compression ratio (communication efficiency), high anti-noise ability, and overall interpretability of the communication workflow.
    \item Ability to be driven by explicit semantic features, selecting appropriate semantic features for compression and generation.
    \item Ability to balance the system's compression ratio and the visual similarity between the transmitted image and the original.
\end{enumerate}  \par
Therefore, by combining semantic communication and large-scale visual generation model techniques, we propose a novel communication paradigm SeFD based on Semantic Feature Decomposition. And under this paradigm, a semantic image communication system TCSCI is proposed. In the TCSCI, the image will be extracted and compressed into natural language descriptions, texture and color semantic features. These features will be further compressed, transmitted over the wireless channel, and reconstructed at the receiver. The reconstructed features will then be input into ControlNet, driving Stable Diffusion model to perform directed semantic feature-based image reconstruction. The main contributions of this paper are summarized as follows.\par

\begin{itemize}
    \item[(1)]
    A novel paradigm called SeFD based on Semantic Feature Decomposition for the integration of semantic communication and large-scale visual generation models is proposed. By decomposing the image into meaningful semantic features at transmitter and explicitly driving the large-scale visual generation model at the receiver to reconstruct the image based on semantic features at the receiver, the paradigm SeFD can achieve an interpretable, controllable, and highly generalizable communication framework.
\end{itemize}
\begin{itemize}
    \item[(2)]
    Based on the paradigm, a novel image semantic communication system called TCSCI is proposed, where the image is composed into natural language descriptions, texture, and color semantic features for transmission, and at the receiver, ControlNet is used to drive a StableDiffusion model for directed semantic feature-based image reconstruction, enabling highly compressed image transmission with high visual similarity, and the system also exhibits good noise resistance.
\end{itemize}
\begin{itemize}
    \item[(3)]
    Experiments demonstrate that the TCSCI can achieve highly visually similar image transmission at extremely low compression rates, outperforming traditional image communication systems and existing semantic communication systems. The TCSCI also exhibits better noise resilience, and can remain robust even under lower signal-to-noise ratio channel conditions.
\end{itemize} \par
This paper is arranged as follows. In section II, the proposed paradigm for the integration of semantic communication and large-scale visual generation models is introduced. In section III, the complete architecture and overall workflow of the TCSCI are introduced in detail. In section IV, the experimental data and visual examples from various experiments are provided to demonstrate the performance of the TCSCI. Finally, conclusions of this paper are drawn in section V.\par

\begin{figure*}[ht]
    \centering
    \includegraphics[scale=0.43]{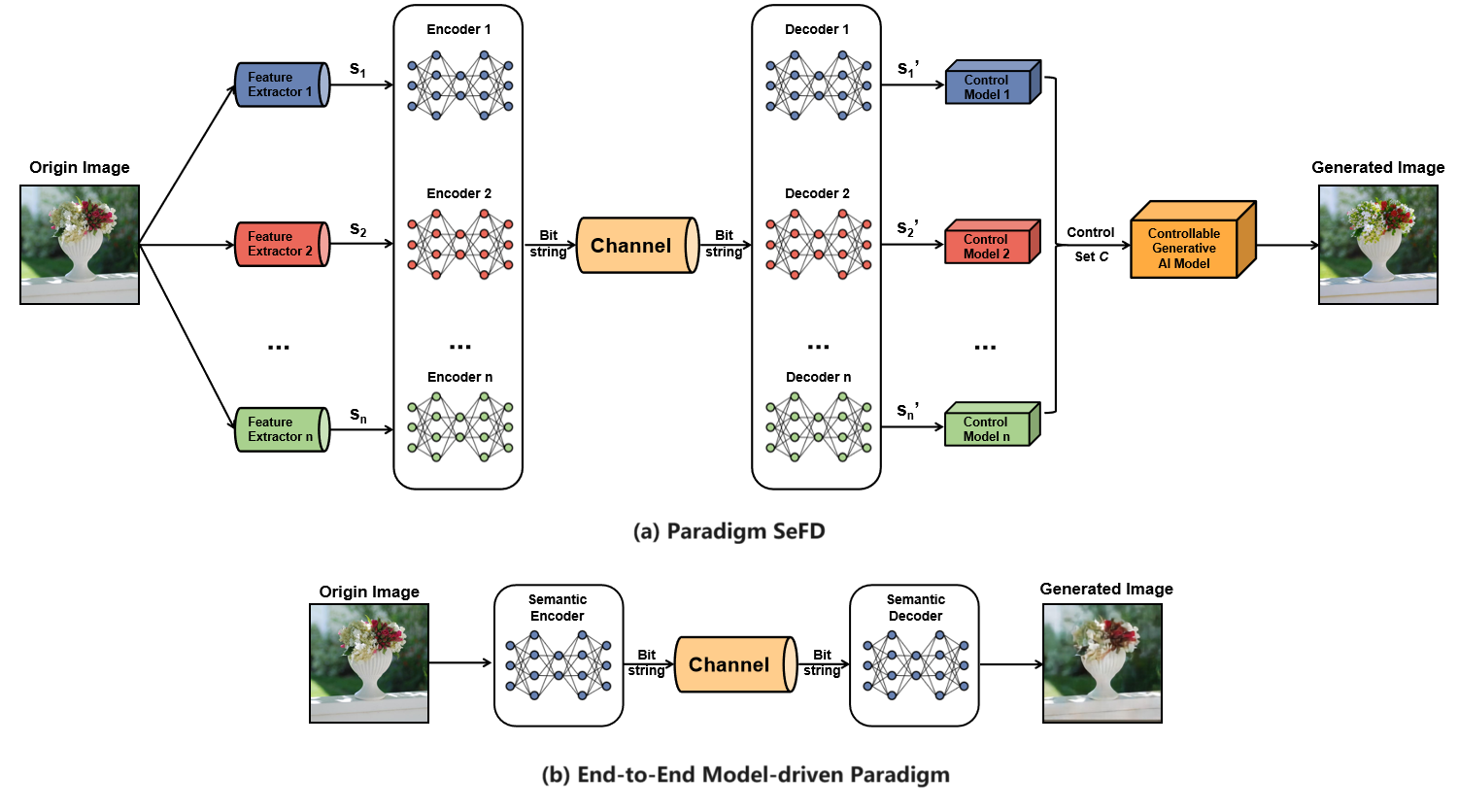}
    \caption{The architecture of the proposed paradigm and the end-to-end model-driven image semantic communication paradigm. In the end-to-end model-driven paradigm, the image undergoes semantic feature extraction and reconstruction directly through a neural network architecture of an autoencoder. The extracted semantic features are based on the black-box-like neural network computation, and the entire process lacks interpretability, editability, and task generalization ability. Whereas in the paradigm we propose, the image is decomposed into different meaningful semantic features, which are then separately encoded, transmitted, and decoded, and at the receiver, they are used to control a large-scale visual generation model to generate the image with specific semantic features.}
\end{figure*}

\section{Paradigm for the Integration of Semantic Communication and Large-scale Visual Generation Models}

As shown in Fig. 1, the existing widely studied model-driven image semantic communication systems, such as \cite{lsci, sci2, stsci, jscc}, can be simplified into a structure of an autoencoder, as shown in sub-figure (b). The image is fed into a semantic encoder, which generates a semantic encoding. This encoding is then transmitted through the channel and reconstructed into an image using a semantic decoder at the receiver. The entire system's “understanding” of the image's semantic information is completely based on the end-to-end trained model structure and parameters, making it difficult to determine whether its understanding and focus on semantic information align with human perception or task requirements. Meanwhile, in this process, the semantic encoding computation is based on the black-box nature of the neural network parameters, lacking controllability and editability, and the entire process lacks interpretability.\par
Despite the fact that through complex model structures and training techniques, image communication systems under this paradigm can achieve decent performance in terms of visual similarity, noise resilience, and compression rate, we still want to obtain an image semantic communication paradigm with strong interpretability and controllability, in order to meet the requirements of different tasks for the fidelity of different semantic information in practical applications. Therefore, a novel paradigm SeFD for the integration of semantic communication and large-scale visual generation models is proposed. As shown in Fig. 1, in our paradigm, the image will first be decomposed into different meaningful semantic features based on human perception or task requirements, such as edge information, depth cues, texture information, etc. At the receiver, these semantic features will then be used to control large-scale visual generation models through advanced AI technologies like ControlNet, to achieve image generation with specific semantic features. The transmission of the semantic features is similar to the end-to-end model-driven paradigm, where they are further compressed, transmitted, and reconstructed through an auto-encoder architecture.\par
The advantages of the proposed paradigm are as follows:

\begin{enumerate}
    \item \textbf{Higher compression ratio.} During the semantic feature decomposition, task-irrelevant information in the image has been largely eliminated. At the same time, the powerful generative abilities of the large model can further help improve the compression ratio of the system. 
    \item \textbf{Stronger anti-noise ability.} In this paradigm, not only can JSCC technology be introduced in the autoencoder architecture for semantic feature transmission, but the large-scale visual generation model itself also has strong anti-noise ability. The combined effect of the two greatly enhances the anti-noise performance of this paradigm.
    \item \textbf{Stronger interpretability.} In this paradigm, the high-level semantic information is human-selected and visualizable, which significantly improves the interpretability compared to the end-to-end model-driven paradigm.
    \item \textbf{Stronger generalization ability.} The decomposition of the semantic features, the transmission of semantic features, and the generation of the images are all decoupled. It is simpler to replace the selected semantic features for different tasks or to change the semantic feature transmission module according to channel conditions, resulting in stronger generalization ability.   
    \item \textbf{Controllability and editability.} It can fully utilize the abilities of the large-scale visual generation model and its control model. The semantic features used for compression and generation in the entire communication process can be manually edited, and the generated images can conform to the editing effects. This makes the more integration of AIGC and communication possible.
\end{enumerate}  \par

\begin{figure*}[ht]
    \centering
    \includegraphics[scale=0.45]{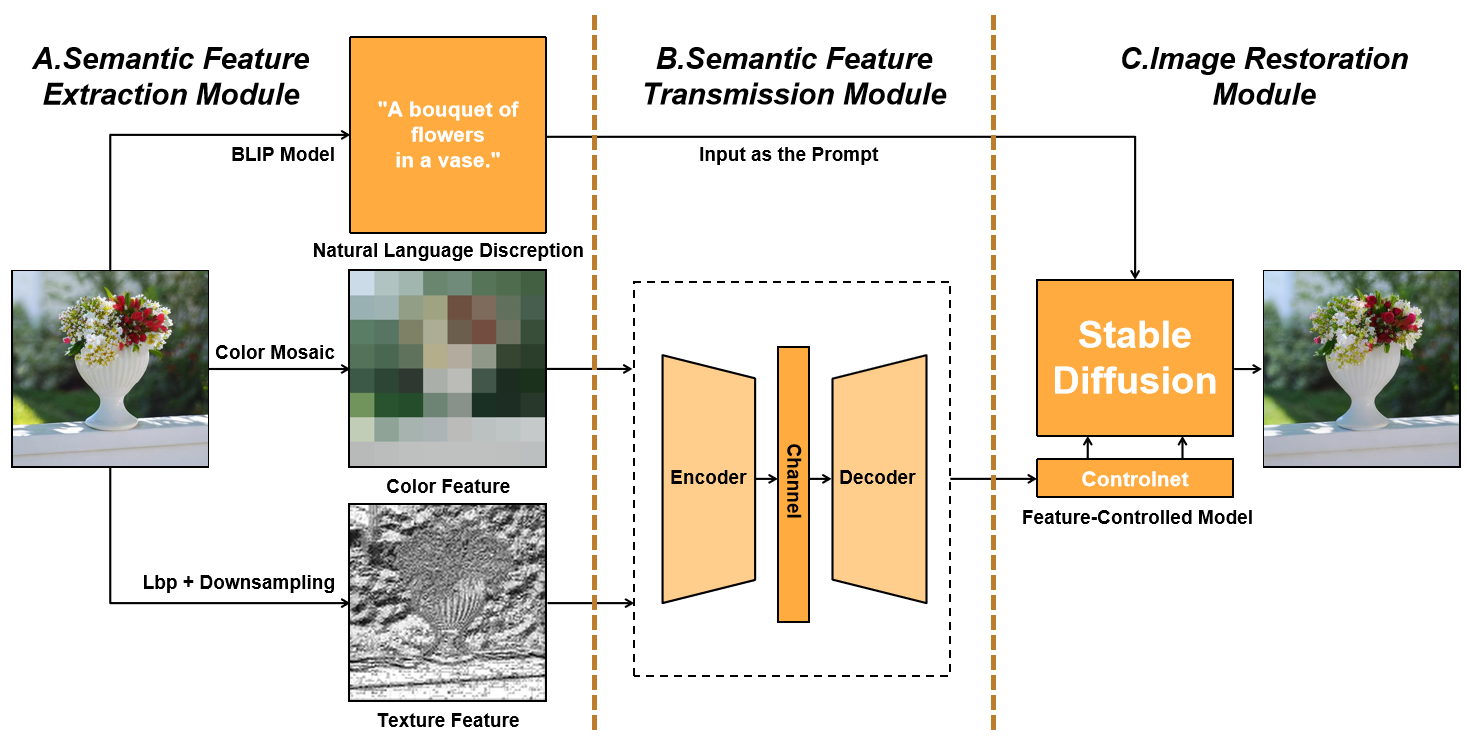}
    \caption{The overall system architecture of TCSCI. The architecture can be divided into three parts based on the communication workflow: A. Semantic Feature Extraction Module, B. Semantic Feature Transmission Module, C. Image Restoration Module. In module A, the image is decomposed into natural language descriptions, color semantic features, and texture semantic features through the BLIP model, downsampling, and LBP algorithm. These features are further compressed in module B, transmitted through the wireless channel, and restored at the receiver. The restored features in module C control the ControlNet to drive the Stable Diffusion for specific semantic-controlled image generation. By overlaying the color and texture features, high visual similarity images are obtained to complete the entire transmission process.}
\end{figure*} 

\section{Semantic Communication System of Images based on Paradigm SeFD}
Based on the paradigm SeFD, specific semantic communication systems can be proposed. Therefore, a Texture-Color based Semantic Communication system of Image (TCSCI) is proposed to transmit image with high visual similarity under extreme compression (bpp < 0.1). Meanwhile, when bpp > 0.1, new semantic feature information will be introduced to improve the system's performance at higher bpp, also showcasing the scalability of the system.\par
The overall system architecture of TCSCI is shown in Figure 2. According to the communication workflow, the system can be divided into three modules: A. Semantic Feature Extraction Module, B. Semantic Feature Transmission Module, C. Image Restoration Module. \par
In module A, the image \mbox{$I$} to be transmitted is decomposed into natural language descriptions, color semantic feature maps, and texture semantic feature maps. The natural language descriptions are computed using the BLIP model \cite{blip}. The color semantic feature maps are obtained by heavily downsampling the image to create a color mosaic. The texture semantic feature maps are computed using the LBP texture algorithm \cite{lbp} to obtain the texture grayscale image, which is then further downsampled to omit some details.
\begin{equation}\label{eqn-1} 
    \mathcal{S} = \{ s_{text}, s_{texture}, s_{color} \},
\end{equation}
where \mbox{$\mathcal{S}$} is the set of semantic features extracted from the input image, the extraction methods will be introduced in detail in subsection A. \par
In module B, the semantic feature maps obtained from module A are further compressed, transmitted through the wireless channel, and then restored with high similarity at the receiver. The architecture of module B is based on existing end-to-end image semantic communication systems \cite{lsci}, serving two purposes: (1) Further compressing the semantic feature maps to improve the overall compression ratio of the system. (2) Introducing joint source-channel coding techniques to reduce the interference of channel noise on the semantic feature maps during wireless transmission, thereby enhancing the system's anti-noise ability.\par
\begin{equation}\label{eqn-1} 
    \mathcal{C} = E(\mathcal{S}),
\end{equation}
\begin{equation}\label{eqn-1} 
    \mathcal{C'} = Channel(\mathcal{C}),
\end{equation}
\begin{equation}\label{eqn-1} 
    \mathcal{S'} = D(\mathcal{C'}),
\end{equation}
where \mbox{$E(\cdot)$} and \mbox{$D(\cdot)$} respectively represent the whole encoding and decoding process in the Module B, \mbox{$C$} represents the coding set transmitted through the wireless channel. \par
In module C, the restored semantic feature maps are used to control the ControlNet, which in turn drives the Stable Diffusion model to generate images with targeted semantic features. By overlaying the color and texture feature maps, and combining them with the natural language descriptions to control the main semantic expression, the Stable Diffusion model generates images that have high visual similarity to the original image,
\begin{equation}\label{eqn-1} 
    I' = SD(s_{text}', Contri(s_{texture}', s_{color}')),
\end{equation}
where \mbox{$SD(\cdot)$} and \mbox{$Contri(\cdot)$} represent the Stable Diffusion and ControlNet, \mbox{$s'$} represents the corresponding semantic features transmitted through the channel and \mbox{$I'$} represents the final generated image.\par
The design rationale and detailed workflow of each module will be introduced in three subsections as follows:\par

\subsection{Semantic Feature Extraction Module}
Inspired by the algorithm in the paper Text-Sketch \cite{textsketch}, we decide to extract vivid visual features from the image to be transmitted as semantic information, and then restore the image at the receiver  using this information. Since the selected visual features will directly affect the compression ratio and the visual similarity of the transmission, it is the most important part of the entire system design. During the design stage, the semantic features to be extracted should consider the following as much as possible:
\begin{enumerate}
    \item \textbf{Comprehensiveness}: the selected features should be able to cover the main visual and semantic information of the image, so that the receiver can have enough information to restore an image with high visual similarity. For example, the Text-Sketch image compression algorithm selects text descriptions and Sketch edge information as the semantic features for compression, but the two cannot well cover the main visual information of the image, so the decompressed image differs greatly from the original image in color and texture, making it difficult for practical application.
    \item \textbf{Feature Overlap}: when selecting features, the overlap of information between the selected semantic features should be minimized as much as possible, so as to avoid redundancy in image semantic feature extraction and compression, thereby improving the compression ratio and ensuring that the system consumes fewer communication resources in the compression and transmission process. For example, texture features and edge features have a very high degree of overlap, and the latter can even be said to be contained in the former, so there is no need to consider these two features simultaneously when selecting semantic features.
    \item \textbf{Interpretability}: the selected semantic features should have interpretability and conform to human cognition, in order to increase the system's controllability and task generalization. For example, the model-driven image semantic communication system with black box-like neural networks severely lacks interpretability and controllability, which is not in line with the concept of semantic communication, and also makes it difficult to adjust the system architecture in specific usage scenarios and control the transmission process. So interpretability is the primary consideration in selecting semantic features.
\end{enumerate}\par
Based on the above reasons, we have finally chosen natural language description, color semantic features, and texture semantic features as the semantic features involved in the transmission process of this system. From the perspective of comprehensiveness, natural language description can highly summarize the most important semantic content in the image, while color and texture information are the visual features that people perceive most strongly in an image, and can cover the main visual information of the image. From the perspective of feature overlap, the color features form a color mosaic after high-ratio downsampling, which almost no longer contains the edges and texture information of the image, and the texture features are grayscale images without color information, so the two can be regarded as an almost orthogonal feature pair with minimal shared information. Finally, from the perspective of interpretability, whether it is natural language description, color features, or texture features, their characteristics are distinct and conform to human's perceptual habits of images, which have high interpretability and controllability. Therefore, it is reasonable to select these three as the semantic features of the system for extraction, compression, and restoration.\par
~\\
\subsubsection{Neural Language Description}
\begin{figure}[h]
    \centering
    \includegraphics[scale=0.6]{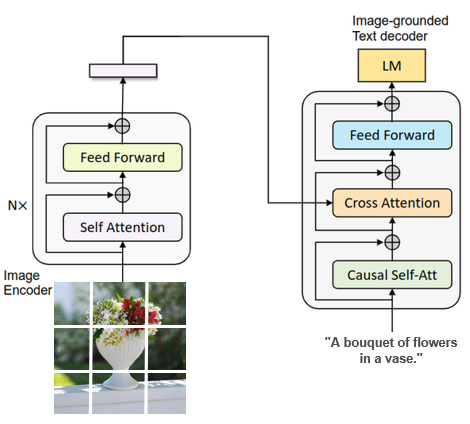}
    \caption{The process of BLIP model. The natural language descriptions are obtained by inputting the image into the BLIP model. In the BLIP model, the Image Encoder divides the input image into patches and encodes them into a series of Image Embeddings. These embeddings are then fed into the Image-grounded Text Decoder to generate the corresponding descriptions.}
\end{figure} 
~\\
\indent The natural language descriptions are computed by the BLIP model, which uses a multimodal mixed structure called MED (Multimodal Mixture of Encoder-Decoder). This structure can establish connections between data of different modalities, enabling effective training on multimodal data and application to multimodal tasks. It is technically mature with excellent performance metrics, and compared to the reverse search prompt technique, the output is more natural language-like and interpretable.\par
As shown in Figure 3, the input image \mbox{$X$} will be fed into the Image Encoder (IE) of the BLIP model, where it will be divided into patches and encoded into a series of Image Embeddings. These embeddings are then input into the Image-grounded Text Decoder (ITD) to generate the corresponding descriptions,
\begin{equation}\label{eqn-1} 
    E_i = IE(X_i),
\end{equation}
\begin{equation}\label{eqn-1} 
    X_{text} = ITD(\mathcal{E}),
\end{equation}
where \mbox{$E_i$} represents the Image Embedding corresponding to the ith patch of the image \mbox{$X$}, \mbox{$\mathcal{E}$} represents the entire set of the Image Embeddings, \mbox{$X_{text}$} represents the natural language descriptions of the image \mbox{$X$}.
~\\
\subsubsection{Color Feature Maps}
~\\
\indent The color feature map is obtained by downsampling the image at a high magnification, resulting in a color mosaic. Before downsampling, the image \mbox{$X$} is first subjected to median filtering,
\begin{equation}\label{eqn-1} 
    col_{ij} = Median ( X_{i'j'} : |i-i'| \le k, |j-j'| \le k ),
\end{equation}
where \mbox{$k$} is the radius of the neighborhood around each pixel, typically set to 3, 5, or 7 pixels. Through median filtering, the pixel value of each pixel is replaced by the median of all the pixel values in its neighborhood. This operation can effectively remove noise in the image while ensuring that the color mosaic values after downsampling are closer to the dominant color tone of their covered region. After median filtering, the image is downsampled according to the downsampling ratio,
\begin{equation}\label{eqn-1} 
    downsamplecol_{ij} = \frac{1}{|\mathcal{D}|} \sum_{(x,y) \in \mathcal{D}} col_{xy}.
\end{equation}
where $\mathcal{D}$ is the downsampling region, and the pixel value of each color mosaic is the average of the pixel values in its representative region.\par
~\\
\begin{figure}[h]
    \centering
    \includegraphics[scale=0.5]{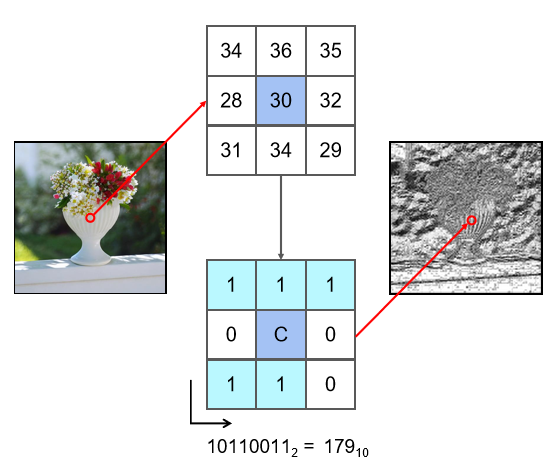}
    \caption{Example of the calculation process of the LBP algorithm. For each pixel in the target image, extract the grayscale values of all pixels in its neighborhood (using a size of 3x3 as an example). As shown in the matrix of the Figure, compare these grayscale values with the grayscale value of the pixel itself. If the value is greater, assign 1; if it is smaller, assign 0. Arrange these 0 and 1 values in a specific order to obtain a binary number. The corresponding decimal value of this binary number represents the grayscale value of the LBP texture feature map at that pixel.}
\end{figure} 
\subsubsection{Texture Feature Maps}
~\\
\indent The texture features of an image describe the texture structure within the image, which can be considered as the local patterns that repeatedly appear in the image. These texture features are an important visual semantic characteristic of the image. The Local Binary Patterns (LBP) algorithm is widely used for extracting texture features from images. The texture features obtained by the LBP algorithm can effectively represent the spatial distribution and shape of the local textures in the image.\par
The basic idea of the LBP algorithm is to compare the grayscale value of each pixel in the image with the grayscale values of its neighboring pixels. Based on this comparison, a binary code is generated, and this binary code represents the texture feature of that pixel. The formula to calculate the LBP value for a pixel with coordinates \mbox{($x, y)$} and a corresponding grayscale value of \mbox{$grey(x,y)$}, with its neighborhood represented as \mbox{$\mathcal{N}$}, is as follows:
\begin{equation}\label{eqn-1} 
    LBP(x,y) = \sum_{(x_k,y_k) \in \mathcal{N} }{\delta (s*g(x_k,y_k)-g(x,y))},
\end{equation}
where \mbox{$\delta$} returns 0 when \mbox{$g(x_k,y_k) \le g(x,y)$}, else returns 1. \mbox{$s$} is a displacement value used to determine the location of the neighboring pixels.\par
As shown in Figure 4, the texture feature maps computed by the LBP algorithm is subjected to a certain degree of downsampling, which sacrifices a portion of the precision in order to improve the compression ratio of the system. By adjusting the downsampling scale of different features, the system can provide different levels of fine-grained control of semantic features in order to accommodate the requirements of various model performance, channel conditions, and transmission targets.

\begin{figure}[ht]
    \centering
    \includegraphics[scale=0.24]{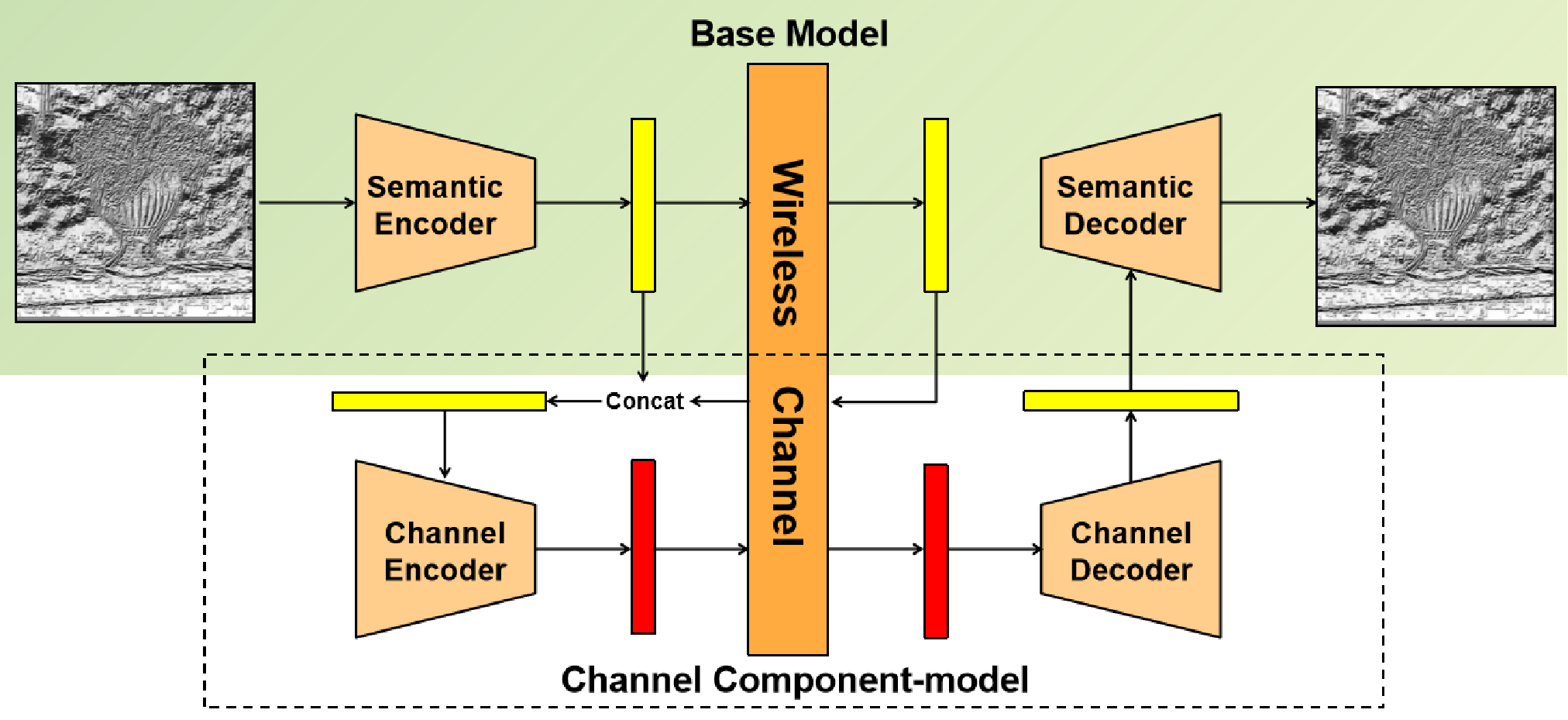}
    \caption{The structure of the Semantic Transmission Module. The semantic transmission module is inspired by the model-driven image semantic system LSCI and consists of two parts: the base model and the channel component-model. In the base model, semantic feature maps are further compressed into semantic latent encoding through convolutional neural networks and are then reconstructed in the semantic decoder. The channel component-model introduces the source-channel joint coding technique to reduce the impact of noise on semantic feature maps during wireless channel transmission, thereby improving the system's anti-noise ability.}
\end{figure} 

\subsection{Semantic Feature Transmission Module}\par
The Semantic Transmission Module aims to compress and transmit the semantic features extracted from Module A and restore them at the receiver. This module is based on the framework presented in reference \cite{lsci} and consists of two sub-models: (1) base model, and (2) channel component-model. The base model is responsible for further compressing the semantic features to improve the system's compression ratio. The channel component-model combines source-channel joint coding techniques to reduce the impact of channel noise and attenuation on the transmission of semantic features, thereby enhancing the system's anti-noise ability. The architecture and workflow of these two sub-models will be introduced separately as follows.\par
\begin{figure}[h]
    \centering
    \includegraphics[scale=0.40]{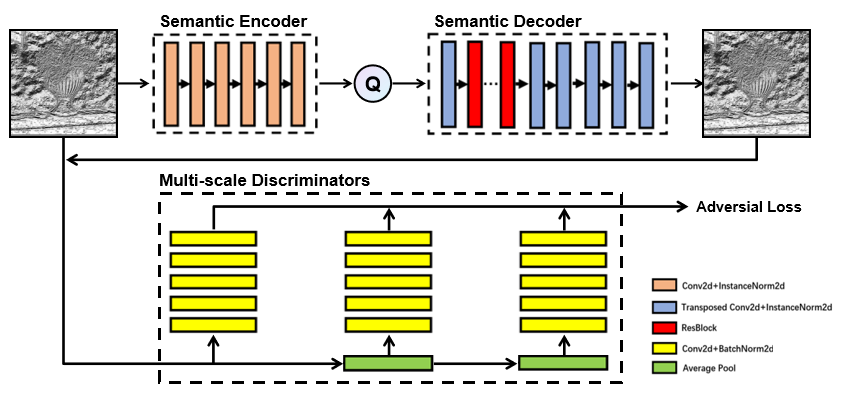}
    \caption{The Base Model includes a Semantic Encoder, a Semantic Decoder, and Multi-scale Discriminators. Where Q represents the nearest neighbor quantization process. With adversarial loss computed by the multi-scale discriminator and the complex model architecture, the semantic decoder can generate semantic feature maps that closely resemble the original ones with high similarity.}
\end{figure} 
\subsubsection{Base Model}
~\\
\indent The base model is used for the compression, quantization, transmission, and decompression of semantic feature maps. According to Figure 6, the architecture of the base model includes a semantic encoder, semantic decoder, and multi-scale discriminators. Q represents the nearest neighbor quantization process. The complex model architecture of the semantic decoder provides it with excellent image restoration capabilities, allowing it to reconstruct semantic feature maps with higher similarity. The multi-scale discriminators are based on the WGAN \cite{gan, wgan} technique and introduce adversarial loss functions to the model, aiming to enhance the quality of the generated images by the semantic decoder.\par

\subsubsection{Channel Component-model}
~\\
\indent Channel interference is a crucial factor that must be considered when designing semantic communication systems. Only by reducing the interference of channel noise on the semantic feature maps can the visual similarity between the generated images and the source images be improved. Traditional channel coding methods suffer from the "cliff effect" problem at low signal-to-noise ratios. Joint source-channel coding (JSCC) techniques can help to alleviate this problem, but JSCC-based systems are only applicable to the specific channel they are trained on, which reduces the generalization ability of JSCC-based systems when deployed in different scenarios. The channel component-models introduced in Module B can serve to ameliorate this problem. Inspired by DeepJSCC-f \cite{jscc-f}, the channel component-models incorporate channel feedback to enhance the noise-resilience of the system. During training, the parameters of the base model are fixed, so when facing different channels, one only needs to replace the channel component-models, which improves the flexibility and applicability of the overall system.\par

\subsection{Image Restoration Module}\par
\begin{figure*}[ht]
    \centering
    \includegraphics[scale=0.9]{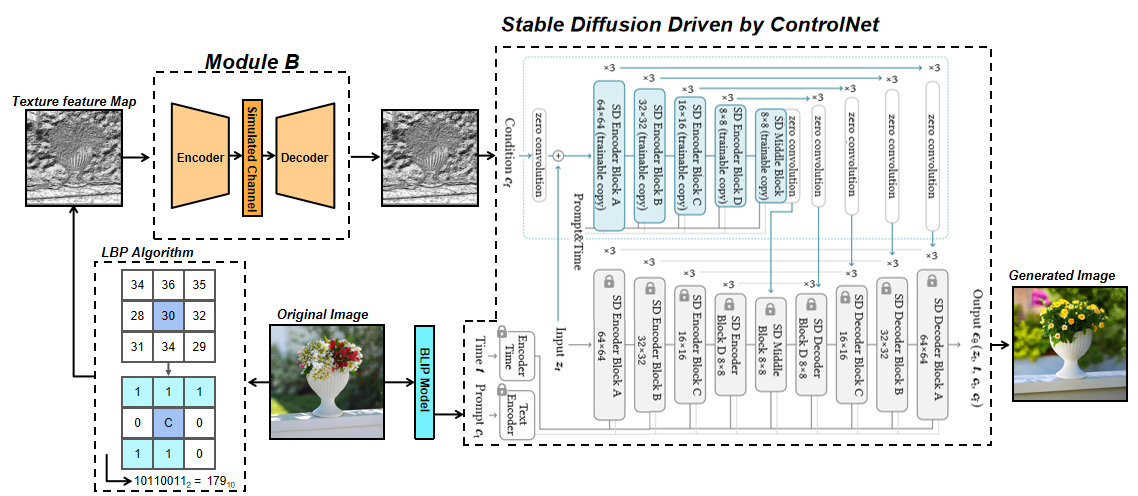}
    \caption{Overall structure and training process of ControlNet. Taking texture feature control as an example, the condition input \mbox{$\textbf{c}_f$} is obtained by processing the original image \mbox{$X$} through the LBP algorithm and Module B of the TCSCI. Concurrently, the prompt input \mbox{$\textbf{c}_t$} is obtained by processing the original image \mbox{$X$} through the BLIP model. With the condition input \mbox{$\textbf{c}_f$} and the prompt input \mbox{$\textbf{c}_t$}, the ControlNet is then trained along with Stable Diffusion to learn to control the generation of images by Stable Diffusion based on the provided texture conditions. During the training process, the parameters of Stable Diffusion are kept fixed, and ControlNet copies the parameters (layers in blue) from Stable Diffusion, inheriting its powerful generative capabilities. Additionally, the zero convolution layers (white layers) are an essential component of ControlNet. ControlNet first applies zero convolution to the control condition, then adds the result to the original input. This combined input then goes through ControlNet's neural network, and the network output undergoes another round of zero convolution before being added back to the original network's output. When ControlNet is untrained, its output is zero, so the added value is also zero. This ensures that the original network's performance is completely preserved, as the ControlNet addition has no effect. Later, when ControlNet is trained, it optimizes the original network, which can be seen as a form of fine-tuning.}
\end{figure*} 
Module C aims to perform image restoration that is highly visually and semantically similar to the original image, by utilizing the semantic features received and restored at the receiver in Module B. This enables the realization of the overall communication flow. This module is primarily composed of the large-scale image generation model Stable Diffusion and the ControlNet.\par 
Stable Diffusion is based on diffusion models \cite{diffusion}, and as one of the more advanced models in the diffusion model family, it employs a more stable, controllable, and efficient approach to generate high-quality images. It has made significant progress in terms of image quality, speed, and cost-effectiveness. Trained on the Laion-2B-en dataset of 2 billion image-text pairs, Stable Diffusion possesses powerful image generation capabilities and task generalization abilities. At the same time, the emergence of ControlNet technology has enabled the control of Stable Diffusion's image generation process using specific semantic features as conditions, addressing the issue of uncontrollable content generation in Stable Diffusion. These factors are the reasons why we ultimately chose Stable Diffusion and ControlNet as the architectural components for Module C.\par
The widely-used ControlNet models currently available include controls for human pose, image edges, image depth, and others. However, the semantic control features involved in these existing works are unable to meet the requirements for high-similarity image semantic communication. Therefore, we have undertaken the training of our own custom ControlNet models for color feature map control and texture feature map control. Figure 7 shows the overall architectural and training process of ControlNet. Taking texture feature control as an example. Training data including texture conditions \mbox{$\textbf{c}_f$} and text conditions \mbox{$\textbf{c}_t$} were first obtained from the original image \mbox{$z_0$} through the LBP algorithm and the BLIP model. It's worth noting that, the \mbox{$c_f$} was transmitted through the simulated channel model in Module B, so that the ControlNet can learn to reduce the impact of information differences caused by channel noise in the transmitted texture feature maps. By combining the noise-resilient ability of the channel component-model as well as the ControlNet, the TCSCI system's overall anti-noise ability will exceed that of a typical image semantic communication system. \par
After the training data preparation, the ControlNet was trained to drive the Stable Diffusion model to perform directed semantic feature-controlled image generation. During the training process, the parameters of Stable Diffusion are kept fixed, and ControlNet copies the parameters from Stable Diffusion, inheriting its powerful generative capabilities. Additionally, the zero convolution layers are an essential component of ControlNet. ControlNet first applies zero convolution to the control condition, then adds the result to the original input. This combined input then goes through ControlNet’s neural network, and the network output undergoes another round of zero convolution before being added back to the original network’s output. When ControlNet is untrained, its output is zero, so the added value is also zero. This ensures that the original network’s performance is completely preserved, as the ControlNet addition has no effect. Later, when ControlNet is trained, it optimizes the original network, which can be seen as a form of fine-tuning. And the training loss is calculated as follows,\par  
\begin{equation}\label{eqn-1} 
    \mathcal{L} = \mathbb{E}_{z_0, \textbf{t}, \textbf{c}_t, \textbf{c}_f, \epsilon \in \mathbb{N}(0,1)}[||\epsilon - \epsilon_{\theta}(z_t, \textbf{t}, \textbf{c}_t, \textbf{c}_f)||_{2}^{2}],
\end{equation}
where \mbox{$z_t$} is produced by adding noise to the input image \mbox{$z_0$} through image diffusion algorithms, \mbox{$t$} represents the number of times noises is added. \mbox{$\epsilon_{\theta}$} represents the model learned to predict the noise added to the noisy image \mbox{$z_t$}. Where \mbox{$\mathcal{L}$} is the overall learning objective of the entire model. This learning objective is directly used in finetuning Stable Diffusion with ControlNet.\par
The overall workflow of Module C is as follows, the color feature maps and texture feature maps extracted in Module A are transmitted to the receiver through Module B and then restored. At the receiver, the color control network and texture control network are fed these features to respectively obtain the color control information and texture control information. These two control signals are then combined to form composite color and texture control information. This is then input, along with the natural language description of the image \mbox{$X$} obtained from Module A, into the Stable Diffusion. Leveraging Stable Diffusion's powerful feature-controllable image generation capabilities, the final image \mbox{$X'$} is then generated.\par
Let \mbox{$SD(\cdot ; \theta)$} represents the trained Stable Diffusion, \mbox{$C(\cdot ; \theta _c)$} and \mbox{$C(\cdot ; \theta_t)$} respectively represents the ControlNet trained with color feature maps and texture feature maps.
\begin{equation}\label{eqn-1} 
    X' = SD(\textbf{c}_{text}; \theta) + C(\textbf{c}_{text}, \textbf{c}_{col}; \theta _c) + C(\textbf{c}_{text}, \textbf{c}_{texture}; \theta _t),
\end{equation}
where \mbox{$\textbf{c}_{text}$}, \mbox{$\textbf{c}_{col}$} and \mbox{$\textbf{c}_{texture}$} respectively represents the natural language descriptions, color feature maps and texture feature maps obtained from the original image \mbox{$X$} through Module A and Module B. 

\subsection{System Extensibility When bpp > 0.1}\par
The TCSCI system is primarily designed for image transmission under extreme compression (bpp < 0.1), as existing image transmission systems struggle to achieve high-similarity image transfer at such low bit rates. When performing extreme compression, due to the limited resources, color and texture features are selected as the semantic features for transmission, as they can cover the main visual and semantic information of the image. This allows the system to achieve optimal results in terms of visual and semantic similarity for the transmitted images. \par
However, TCSCI is not limited only to the extreme compression domain. When bpp > 0.1, within the paradigm SeFD, more semantic features can be introduced. These features are decomposed at the transmitter and then used to control the image generation at the receiver. For example, at bpp = 1.5, depth information calculated using MiDAS \cite{midas} can be introduced in addition to color and texture features, and the combination of these features will result in the optimal generation quality. \par
As the bpp continues to increase, the system can introduce new semantic features to achieve better transmission performance. However, the selection of features and the issue of feature overlap cannot be ignored in this process, but these are not within the scope of the current discussion.\par

\section{Experiment Results}
This section is mainly introduced the relevant testing settings, including the dataset for TCSCI's test, the introduction of baseline and the image similarity metrics used for system's performance evaluation. Discription and figures are given to show how the TCSCI surpass the traditional image communication systems and existing semantic image communication at extremely low bpps or lower signal-to-noise ratios at visual or semantic level.\par
\begin{figure*}[ht]
    \centering
    \includegraphics[scale=0.5]{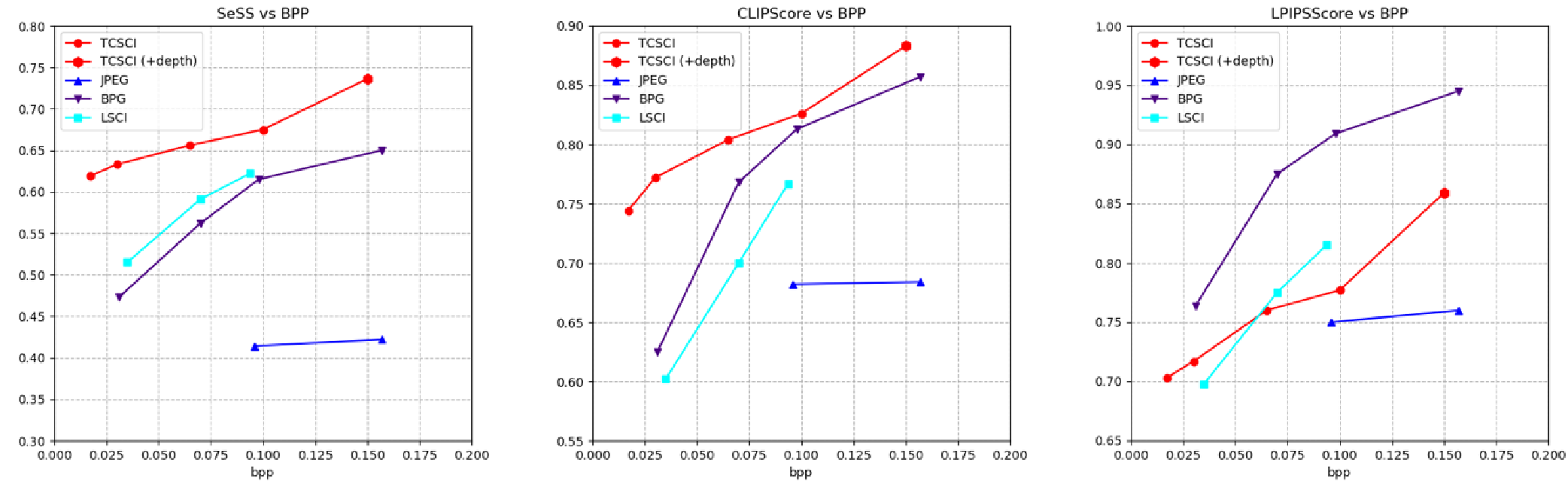}
    \caption{Comparisons between the semantic communication system TCSCI, LSCI and the traditional image compression algorithms JPEG and BPG at different bpps. In this experiment, the channel is assumed to be lossless, so the information redundancy brought by channel coding in actual transmission for JPEG and BPG is not calculated. It can be found that in the extreme compression domain, TCSCI outperforms the existing traditional image compression algorithms and the semantic communication system LSCI in terms of the more semantically-oriented metrics SeSS and CLIPScore. And in the more structure-oriented metric LPIPSScore, TCSCI is still superior to the JPEG algorithm, and exceeds the semantic communication system LSCI at lower bpps. In addition, when bpp > 0.1, depth feature is introduced into TCSCI to achieve better transmission performance.}
\end{figure*} 
\begin{figure*}[ht]
    \centering
    \includegraphics[scale=0.45]{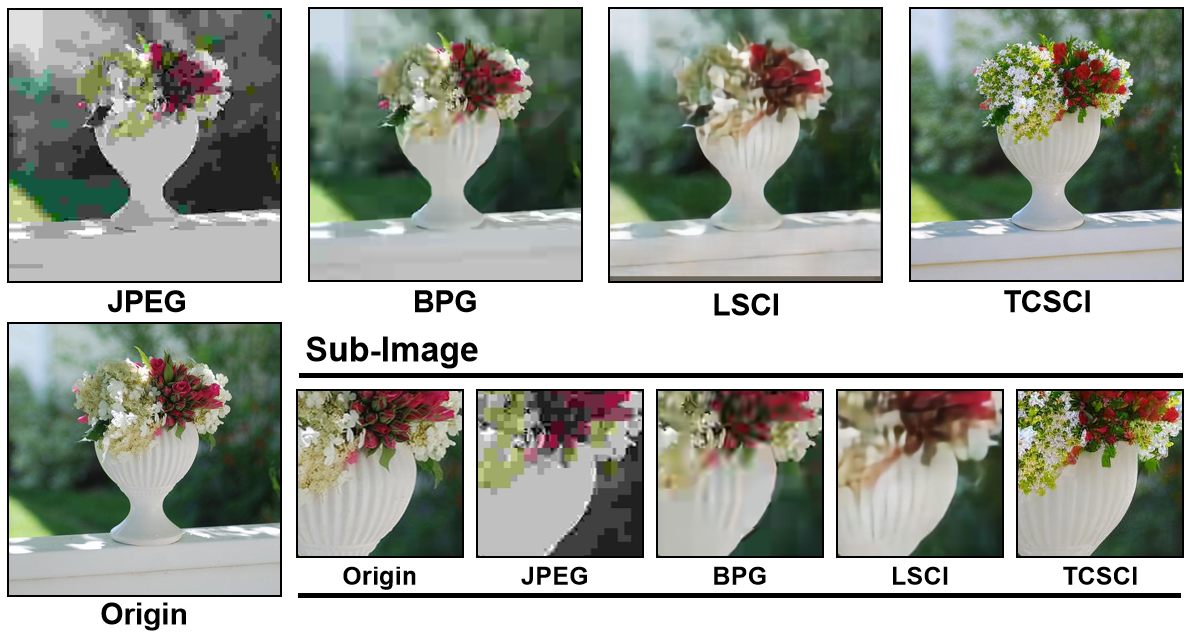}
    \caption{Visual example of the image compressed by image semantic communication system TCSCI, LSCI and image compression algorithm JPEG and BPG at extremely low bpp around 0.1. It can be seen that when bpp is around 0.1, JPEG has a large amount of distortion and deformation, losing most of the color and texture details, while the quality of BPG is relatively better, but the image blurring is unavoidable. These phenomena will hinder human semantic perception of the image. LSCI, on the other hand, is limited by the low compression ratio and cannot achieve a high-quality end-to-end image restoration. Although TCSCI has some differences in details compared to the original image, it has a high similarity to the original in terms of overall semantic representation and visual effects, and also has a high image quality without distortion and blurring.}
\end{figure*} 

\subsection{Experiment Settings}\par
The experimental data is sampled from the COCO2017 dataset. The baseline of the experiments is chosen both in traditional and semantic communication algorithms and systems. Image compression algorithms JPEG and BPG is involved. JPEG is a widely used image compression standard that is commonly employed in image transmission and storage. On the other hand, as an emerging image compression algorithm, BPG adopts more advanced encoding techniques, allowing it to maintain better visual details and image clarity than JPEG at the same compression rate. Semantic image communication system LSCI is also used for comparison, which is a novel generalized semantic communication system. \par
Given the extreme compression domain, pixel-based metrics like PSNR and structure-based metric like SSIM is not suitable for measuring visual and semantic similarity between images effectively. To better evaluate the performance of the TCSCI, three image similarity metrics are used for system's performance evaluation, including LPIPSScore, CLIPScore \cite{clipscore} and SeSS \cite{sess}. The LPIPSScore is calculated by taking 1 minus the perceptual loss between the two images. Due to the constraints of the convolutional layer architecture, LPIPSScore still considers a lot of low-level image information, rather than focusing more on the differences in semantic-level information. Therefore, in this metric, the metric score of TCSCI may be relatively lower, but this also reflects that the design of TCSCI is different from the more non-semantic-aware model-driven image communication system, as it focuses more on the information at semantic level.\par 
CLIPScore calculates the similarity between images based on the multi-modal CLIP model \cite{clip}, which has the ability to ability to capture the global semantic representation of images. SeSS is based on the Segment Anything Model \cite{sam} and Scene Graph Generation technology. By transforming the calculation of differences between images into the calculation of differences between the object-relation networks represented by the images, SeSS can "understand" the semantic-level representation of the images and more accurately measure differences between the images at semantic level.\par
To fully demonstrate the performance of TCSCI, three experiments were carried out, corresponding to tests of the system's communication efficiency, noise resistance, and semantic controllability.\par

\begin{figure*}[ht]
    \centering
    \includegraphics[scale=0.5]{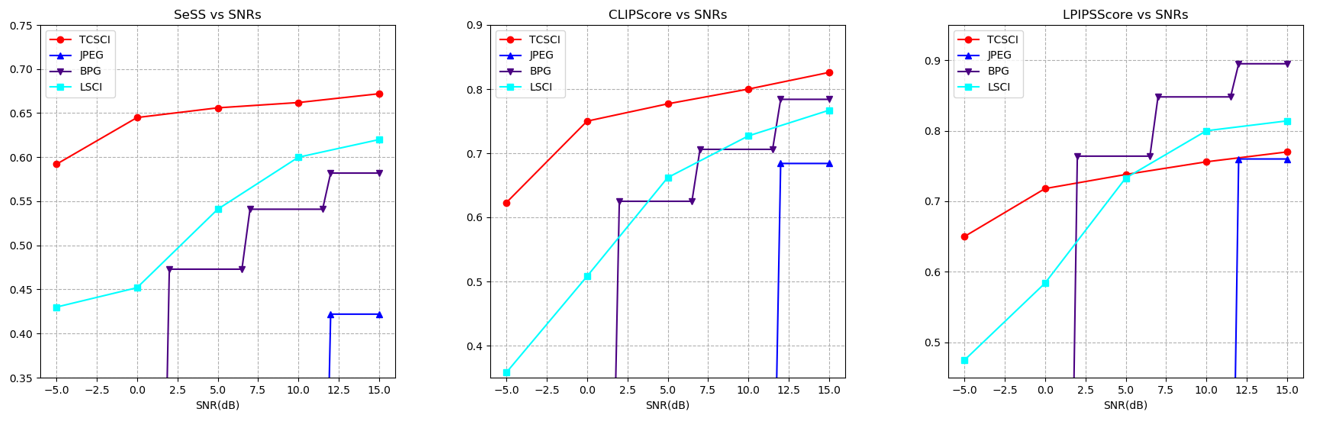}
    \caption{Comparisons between the semantic communication system TCSCI, LSCI and the traditional image compression algorithms JPEG and BPG under different SNRs. In the experiments, both JPEG and BPG employed 1/2 LDPC channel coding. In terms of modulation, TCSCI and LSCI adopted 16QAM modulation, while JPEG and BPG varied the modulation scheme with SNR, using BPSK, QPSK, 16QAM, and 64QAM in succession. The experiments ensured that the \mbox{$\frac{Number \, of \, Transmitted \, Bits}{Bits \, per \, Symbol}$} is equal for all the systems. It can be found that on the more semantically-focused metrics of SeSS and CLIPScore, TCSCI surpasses JPEG, BPG, and LSCI. And on the more structurally-focused metric of LPIPSScore, TCSCI also outperforms the JPEG algorithm, and under lower SNR conditions, it even exceeds LSCI and BPG. Meanwhile, to a certain extent, compared to other systems, the image metrics of TCSCI degrade more slowly as the signal-to-noise ratio decreases. These results all demonstrate the powerful anti-noise ability of TCSCI.}
\end{figure*} 

\begin{figure}[ht]
    \centering
    \includegraphics[scale=0.22]{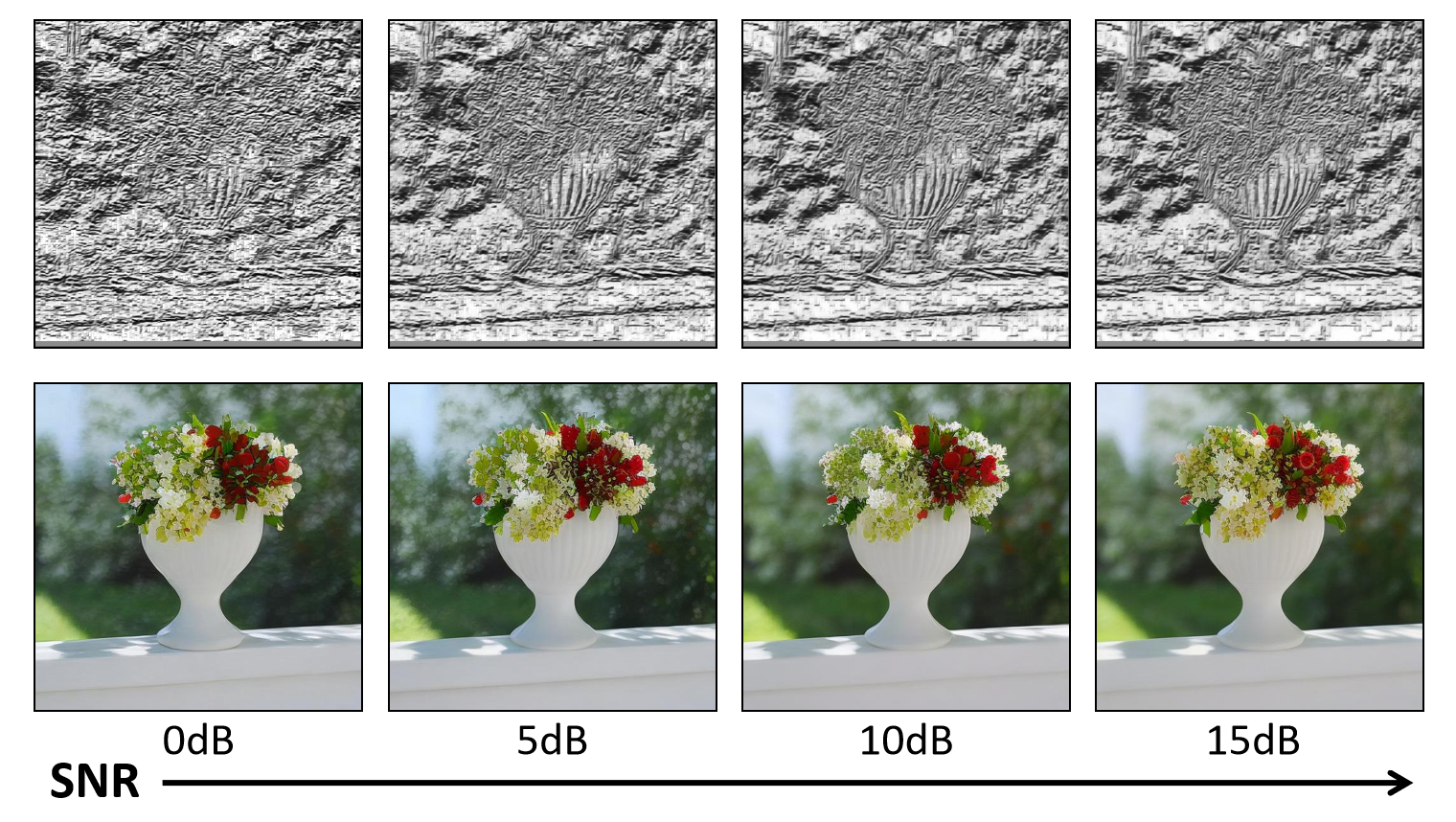}
    \caption{Visual examples of images transmitted by TCSCI under different SNRs. The top row shows the results of transmitting the texture feature maps through Module B under different signal-to-noise ratios, and the bottom row shows the corresponding TCSCI system outputs under those SNR conditions. It can be observed that within a certain range, the semantic and visual changes in the transmitted images are relatively small, maintaining an extremely high visual similarity to the original image. Through the joint effects of the channel component-model in Module B and the ControlNet trained on noisy data in Module C, TCSCI exhibits good noise resistance.}
\end{figure} 

\begin{figure*}[ht]
    \centering
    \includegraphics[scale=0.65]{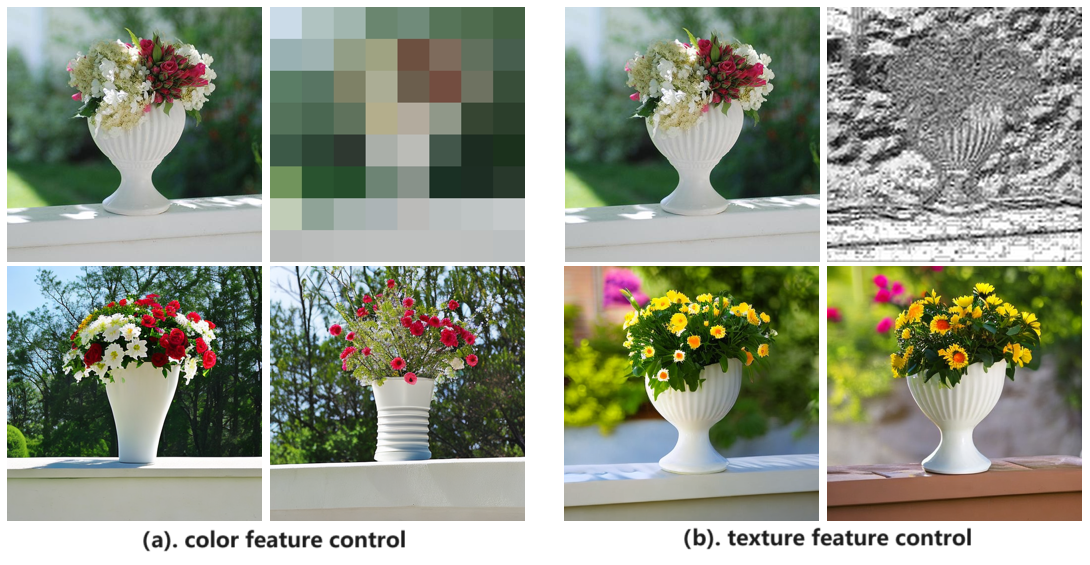}
    \caption{Visual examples of images controlled separately by color control process and texture control process of the TCSCI. Each subfigure follows the same layout - the top-left is the original image, the top-right shows the semantic feature maps, and the bottom portion displays the generated images. It can be seen, such controlled image generation produces visible and controllable results, which proves the decoupled and scalable nature of the semantic feature control process in TCSCI. This also showcases the potential of TCSCI to flexibly substitute the controlled semantic features in order to adapt to different task requirements.}
\end{figure*} 
\begin{figure*}[ht]
    \centering
    \includegraphics[scale=0.4]{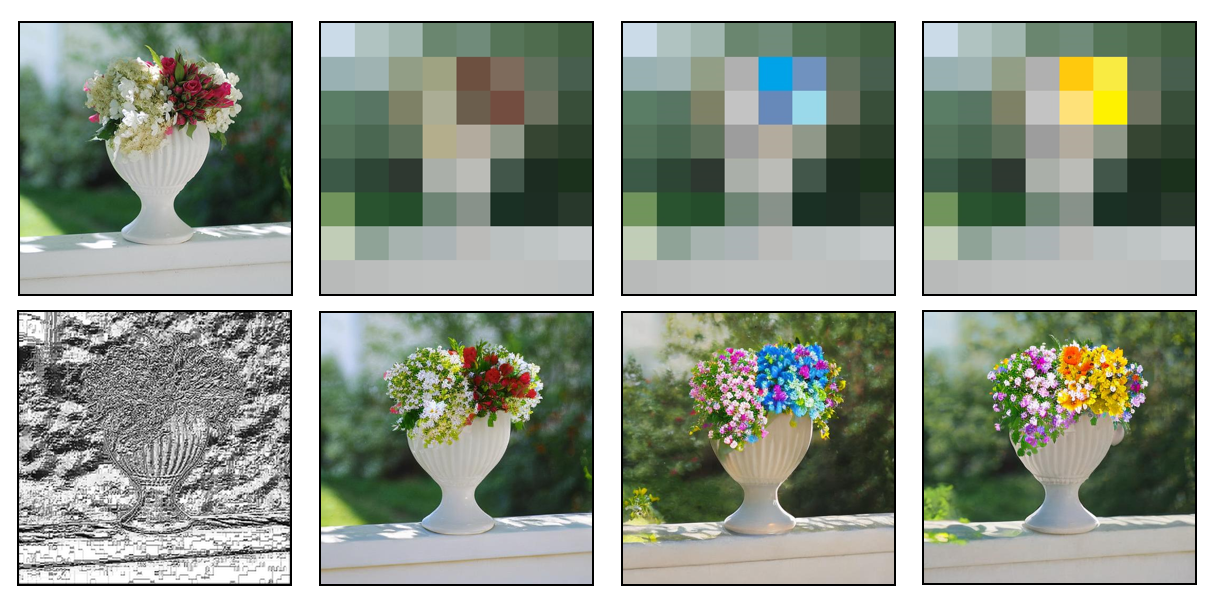}
    \caption{Visual examples to demonstrate the semantic editability of TCSCI. The top-left subfigure is the original image, the bottom-left is the texture feature maps, the others are edited color feature maps and the corresponding output images of TCSCI. Using the example of color semantic features, it can be observed that the TCSCI system can achieve directed semantic control of the output image by modifying the semantic features prior to generating the final image at the receiver.}
\end{figure*} 

\subsection{Compression Performance}\par
Figure 8 shows the comparison between semantic communication system TCSCI, LSCI and traditional image compression algorithm JPEG and BPG. In this test, the impact of the channel is not considered, so the redundancy introduced by channel coding in actual transmission is also not taken into account in the bpp calculation for JPEG and BPG, while the compression limits of the JPEG and BPG algorithms are approximately around bpp=0.1 and bpp=0.03 respectively.\par
It can be found that in the extreme compression domain, TCSCI outperforms the existing traditional image compression algorithms and the semantic communication system LSCI in terms of the metrics SeSS and CLIPScore. Both metrics more focused on information of images at semantic level, which means the TCSCI can better preserve the semantic information of the image under extremely low compression rates. 
And in the more structure-oriented metric LPIPSScore, TCSCI is still superior to the JPEG algorithm, and exceeds the semantic communication system LSCI at lower bpps. And in the more structure-oriented metric LPIPSScore, TCSCI is still superior to the JPEG algorithm, and exceeds the semantic communication system LSCI at lower bpps.\par
In addition, when bpp > 0.1, depth feature is introduced into TCSCI to achieve better transmission performance. After introducing depth information, TCSCI (+depth) achieved relatively high metrics around 0.15 bpp. This indicates that as a concrete instance of the Paradigm SeFD, TCSCI is not limited to the extremely compressed domain (bpp < 0.1), but has sufficient flexibility and extensibility. As the bpp increases, TCSCI is able to improve transmission performance by reasonably introducing more semantic features. \par
In Figure 9, visual examples are given to showcase how TCSCI outperforms the traditional image compression algorithms and model-driven semantic communication system LSCI in terms of semantic and visual aspects when performaning extremely low bpp compression. All the images are compressed by different algorithms or systems at bpp around 0.1. The image compressed by JPEG exhibits significant distortion and deformation, as well as loss of color and texture details. While BPG has better compression quality, the image still exhibits substantial blurring. These will impact human semantic perception of the image. At the same time, limited by the extremely low compression rate, LSCI finds it difficult to achieve high-quality end-to-end image restoration, with blurriness and differences still present. Although TCSCI has some differences in details compared to the original image, its overall semantic representation and visual effects are highly similar to the original, and the overall image quality is better, without the presence of distortion and blurring.\par

\subsection{Anti-noise Ability}\par
The anti-noise ability is an important aspect in evaluating image communication systems. We tested the transmission performance of different systems in AWGN channels with various signal-to-noise ratios. The JPEG and BPG algorithms both use 1/2 LDPC channel coding. In the experiment, TCSCI and LSCI both use 16QAM modulation, while JPEG and BPG use BPSK, QPSK, 16QAM, and 64QAM modulation in succession as the signal-to-noise ratio changes, ensuring that the \mbox{$\frac{Number \, of \, Transmitted \, Bits}{Bits \, per \, Symbol}$} is equal for all systems. Since JPEG and BPG have lower bounds on the compression ratio in the extreme compression domain, we only compared the performance of JPEG with 64QAM and BPG with QPSK, 16QAM, and 64QAM modulation.\par
As shown in Figure 10, TCSCI surpasses JPEG, BPG and LSCI in terms of the more semantically-focused metrics SeSS and CLIPScore. And on the more structurally-focused metric of LPIPSScore, TCSCI also outperforms the JPEG algorithm and exceeds LSCI and BPG under lower SNRs condition. Meanwhile, to a certain extent, compared to other systems, the image metrics of TCSCI degrade more slowly as the signal-to-noise ratio decreases. These results all demonstrate the powerful noise resistance ability of TCSCI.\par

The visual examples transmitted by the TCSCI under different SNRs are shown in Figure 11. The top part shows the results of transmitting the texture feature maps through Module B under different SNR conditions, and the bottom part shows the final TCSCI outputs at the corresponding SNRs. Consistent with the data in Figure 10, the TCSCI system, within a certain range of SNR variations, exhibits little change in the semantic representation and visual quality of the transmitted images, maintaining an extremely high visual similarity to the original. \par
From the texture feature maps in the top part, it can be seen that Module B of TCSCI, by introducing the channel component-model with JSCC technologies, effectively protects the texture feature maps. The occasional noise points and discrepancies are also mitigated by the powerful image processing ability of the ControlNet trained in Module C. This is why TCSCI is more robust to noise compared to general image traditional or semantic communication systems.\par

\subsection{Semantic Controllability and Editability}\par

This subsection is primarily focused on showcasing the semantic controllability and editability capabilities of the TCSCI system. Figure 12 demonstrates the effect of separately controlling color and texture in the TCSCI system. Each subfigure follows the same layout - the top-left is the original image, the top-right shows the semantic feature maps, and the bottom portion displays the generated images. It can be observed that using only the texture or color feature maps for image generation produces visible and controllable results. This proves the decoupled and scalable nature of the semantic feature control process in TCSCI. In other words, TCSCI has strong generalization capabilities, allowing it to substitute the controlled semantic features based on different task requirements, in order to achieve better application-specific results.\par

Figure 13 proves the the semantic editability of TCSCI through visual examples. Taking color feature control as an example, TCSCI can achieve directed semantic control of the output image by modifying the semantic features prior to generating the final image at the receiver. Such semantic editability brings more possibilities for more flexible, controllable, and intelligent communication. Users can selectively adjust the semantic attributes of images to achieve more precise task customization. This paradigm holds the promise of inspiring deeper integration between the domains of semantic communication and AIGC fields.\par 

\section{Conclusion}
In this paper, a novel paradigm SeFD based on Semantic Feature Decomposition is proposed. Based on the paradigm SeFD, a image semantic communication system TCSCI is proposed. By combining semantic communication and large-scale visual generation models, TCSCI extremely compress the image into their natural language descriptions, texture and color feature maps. At the receiver, these semantic features are input to ControlNet, which then drives Stable Diffusion to generate images with directed semantic characteristics, thereby achieving semantically-controlled image transmission. Experiments show that TCSCI can achieve highly compressed image transmission with high visual similarity, and the system also exhibits good noise resistance. Meanwhile, as a feasible example of the proposed paradigm SeFD, TCSCI proves that our paradigm possesses the advantages of being interpretable, controllable, and editable. This provides insights and inspiration for the convergence of the semantic communication and AIGC domains.\par
However, TCSCI still faces certain challenges. For instance, the high computational power requirement of the large model Stable Diffusion at the receiver, as well as the high latency in image generation, are issues that cannot be overlooked in practical applications. Additionally, Stable Diffusion's deficiencies in generating human faces and text also limit the application scenarios of TCSCI. Addressing these problems will require continued advancements in both the communication and AI domains.\par

\section{Acknowledgements}
This work is supported by the National Key R\&D Program of China under Grant 2022YFB2902100.\par

\bibliographystyle{IEEEtran}
\bibliography{new}

\end{document}